%% file: main.tex
\documentclass[10pt,twocolumn,letterpaper]{article}

\usepackage[applications]{wacv}  
\usepackage{algorithm}
\usepackage{algpseudocode}
\usepackage{multirow}
\usepackage{siunitx}
\usepackage{comment}
\usepackage[table]{xcolor}
\definecolor{rowblue}{RGB}{232, 242, 255}
\definecolor{rowred}{RGB}{255, 232, 232}


\definecolor{wacvblue}{rgb}{0.21,0.49,0.74}
\usepackage[pagebackref,breaklinks,colorlinks,allcolors=wacvblue]{hyperref}

\def\wacvPaperID{586} 
\def\confName{WACV}
\def\confYear{2027}

\title{Pocket-STVG: lightweight architecture for Spatio-Temporal Video
Grounding}

\author{Alberto Presta$^{1*}$\quad  Michal Byra$^{1,2}$ \quad Grzegorz Stefański$^{1,*}$ \\  Karol Szurkowski$^{1}$ \quad  Eryk Kołodziejczyk$^{1}$ \quad Krzysztof Arendt$^{1}$
 \\
 $^{1}$Samsung AI Center, Warsaw, Poland  \quad  $^{2}$IFTR, Polish Academy of Sciences, Warsaw, Poland \\
{\tt\small $^{*}$corresponding author: a.presta@samsung.com}
}

\begin{document}
\maketitle
\input{sec/0_abstract}    
\input{sec/1_intro}

\input{sec/2_rl}

\input{sec/3_method}

\input{sec/4_exp}

\input{sec/5_concl}

{
    \small
    \bibliographystyle{ieeenat_fullname}
    \bibliography{main}
}

\clearpage
\appendix
\input{sec/7_appendix}

\end{document}

%% file: sec/0_abstract.tex
\begin{abstract}
Spatio-Temporal Video Grounding (STVG) aims to localize the spatio-temporal tube in a video corresponding to a natural language query. While recent methods achieve strong performance in fully supervised, weakly supervised, and zero-shot settings, they typically rely on computationally expensive architectures, complex training pipelines, or multimodal large language models. We present Pocket-STVG (P-STVG), a lightweight cascade architecture that addresses STVG by combining efficient pre-trained components instead of large end-to-end models. P-STVG integrates a temporal-aware video encoder based on MobileViCLIP, a spatial encoder-decoder derived from MDETR, and a shared aligned text encoder. Temporal localization is performed through either a lightweight 1D U-Net or a simple thresholding strategy, enabling the same framework to operate in both weakly supervised and zero-shot settings. Furthermore, video representations are precomputed independently of the query, yielding an indexing-friendly pipeline for efficient inference and large-scale video collections. Despite requiring fewer than 90M parameters, P-STVG performs on par with weakly supervised methods and improves on earlier zero-shot approaches at a fraction of their memory and computational cost, establishing a favorable performance-efficiency trade-off for STVG.


\end{abstract}

%% file: sec/1_intro.tex
\section{Introduction}

\begin{figure}
    \centering
    \includegraphics[width=0.8\columnwidth]{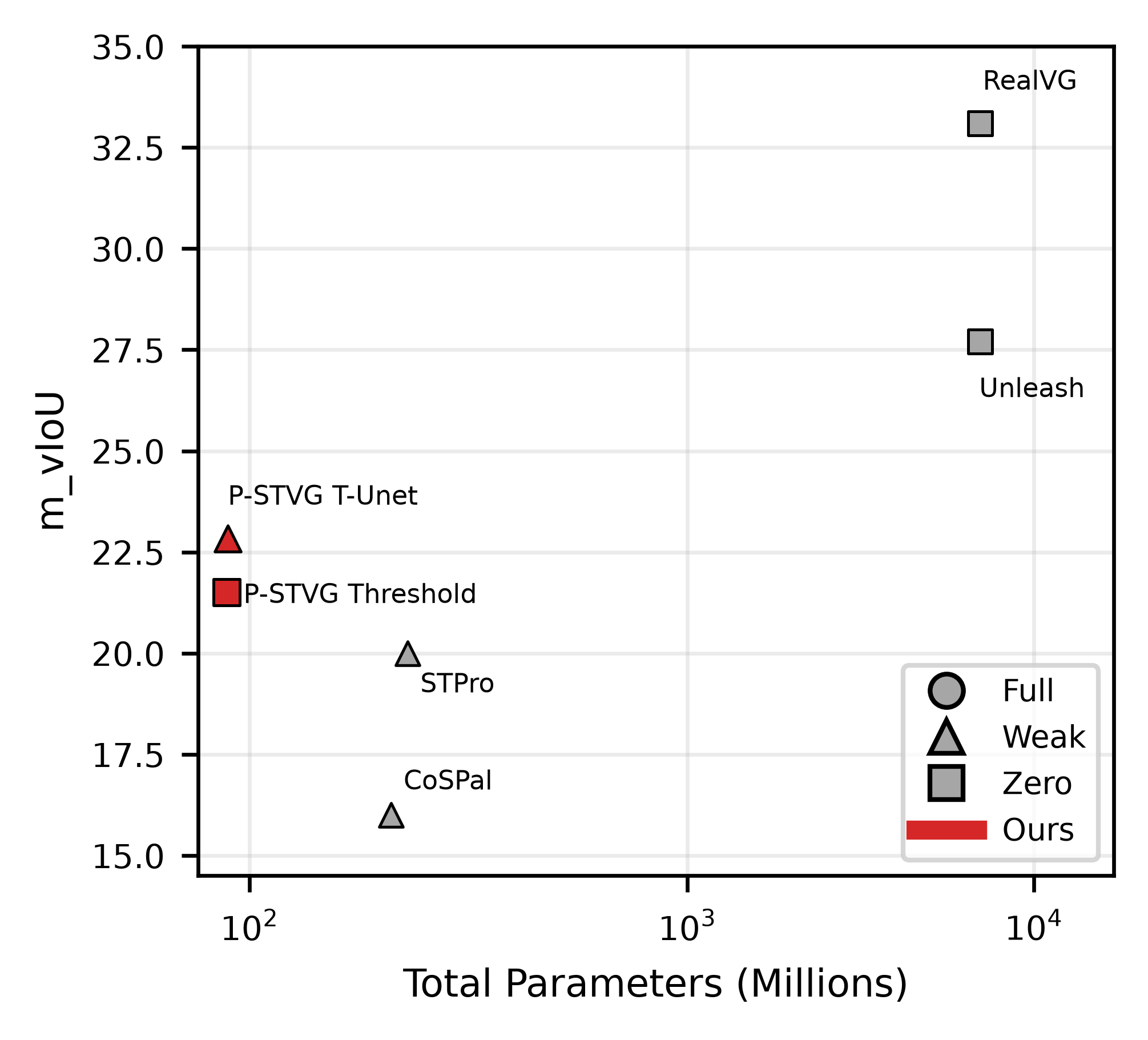}
    \caption{Performance–efficiency comparison on HC-STVG v2. P-STVG occupies a compact operating point, retaining useful grounding accuracy with substantially fewer parameters than recent weakly supervised and zero-shot methods. Marker shapes indicate the supervision setting.}
    \label{Teaser}
\end{figure}

Spatio-Temporal Video Grounding (STVG) focuses on localizing, both spatially and temporally, the target action specified by a textual query within a given video sequence.
The rapid advancement of multimodal architectures and transformer-based models \cite{carion2020end, vaswani2017attention}, along with its importance in key applications such as anomaly detection and robotics \cite{zeng2025futuresightdrive, wang2024facialpulse, gao2025stead}, has propelled STVG into the forefront of research within the deep learning community.
However, it is also challenging task as it requires merging complex information originating from different modalities, discarding noise and focusing on specific important targets.  
While first approaches divided STVG into cascades of sub-stask \cite{zhang2020does,tang2021human, yamaguchi2017spatio}, more recent methods aim to develop compact end-to-end transformer based architectures that output the spatio-temporal tube directly. These models can be categorized into three distinct approaches: fully supervised methods \cite{gu2025knowing, gu2024context, yang2022tubedetr}, which rely on costly frame-level annotations; weakly-supervised methods \cite{liu2024single, garg2025stpro, li2023winner}, which utilize only video-text pairs or partially labeled data; and zero-shot methods \cite{bao2024e3m, yang2025unleashing}, which leverage powerful pre-trained vision-language models \cite{radford2021learning, liu2024grounding, ravi2024sam} to eliminate the dependency on annotated data.
Fully supervised methods have achieved the best results in terms of both temporal and spatial detection, but they require dense data annotations and  often overfit to training dataset, lacking generalization capability.
On the other hand, both weakly and zero-shot approaches rely on pre-existing knowledge coming from huge pre-trained models, achieving remarkable results, even if inferior with respect to fully-supervised methods. However, these approaches often employ highly complex iterative optimization strategies \cite{bao2024e3m} or depend on large-scale models during inference \cite{yang2025unleashing}, which poses challenges for their practical deployment in offline scenarios on resource-constrained edge devices.
In general, existing approaches focus on optimization of the spatio-temporal performance rather than models memory footprint, jeopardizing practical scenarios of using the system on mobile devices.
On the contrary, recently it has been shown that a large image-text foundation models such as CLIP \cite{radford2021learning} can be distilled into  smaller architectures for efficient learning  \cite{vasu2024mobileclip}, with straightforward extension to videos~\cite{yang2025mobileviclip}.

Inspired by recent efficiency-oriented architectures, we propose \textbf{Pocket-STVG} (P-STVG), a lightweight architecture for Spatio-Temporal Video Grounding. The pipeline consists of two stages: indexing and inference. During indexing, spatial and temporal video embeddings are precomputed independently of the query, enabling efficient processing of long videos and large-scale collections. During inference, the query is encoded into text embeddings, after which a lightweight temporal decoder first performs temporal grounding and a spatial decoder subsequently predicts the target bounding boxes. To maximize parameter efficiency, we adopt cross-architecture synthesis by aligning the video-text encoder with the spatial decoder for character-level span prediction, while temporal localization is performed by a lightweight 1D U-Net that effectively models long-range temporal dependencies.

Despite requiring fewer than 90M parameters, P-STVG achieves competitive performance with both weakly supervised and zero-shot methods while substantially reducing memory footprint and computational cost. As illustrated in Fig.~\ref{Teaser}, P-STVG occupies a favorable performance--efficiency operating point, combining compact model size with a simple, non-iterative inference pipeline designed for resource-constrained video grounding scenarios.
The contributions of this work are as follows:

\begin{itemize}
    \item We propose P-STVG, a lightweight and indexing-friendly architecture for STVG with fewer than 90M parameters, designed to balance grounding accuracy and computational efficiency in resource-constrained settings.
    
    \item We introduce a query-independent video indexing stage that precomputes spatial and temporal representations, enabling efficient repeated inference over long videos and large-scale video collections.
    
    \item We achieve competitive spatio-temporal grounding performance with weakly supervised and zero-shot methods while substantially reducing model size, memory footprint, and computational complexity.
    
    \item We introduce a lightweight temporal decoder based on a 1D U-Net, together with a parameter-free thresholding alternative, enabling the same framework to operate under both weakly supervised and zero-shot temporal grounding settings.
\end{itemize}

%% file: sec/2_rl.tex
\begin{figure*}[!t]

    \centering
    \includegraphics[width=0.85\linewidth]{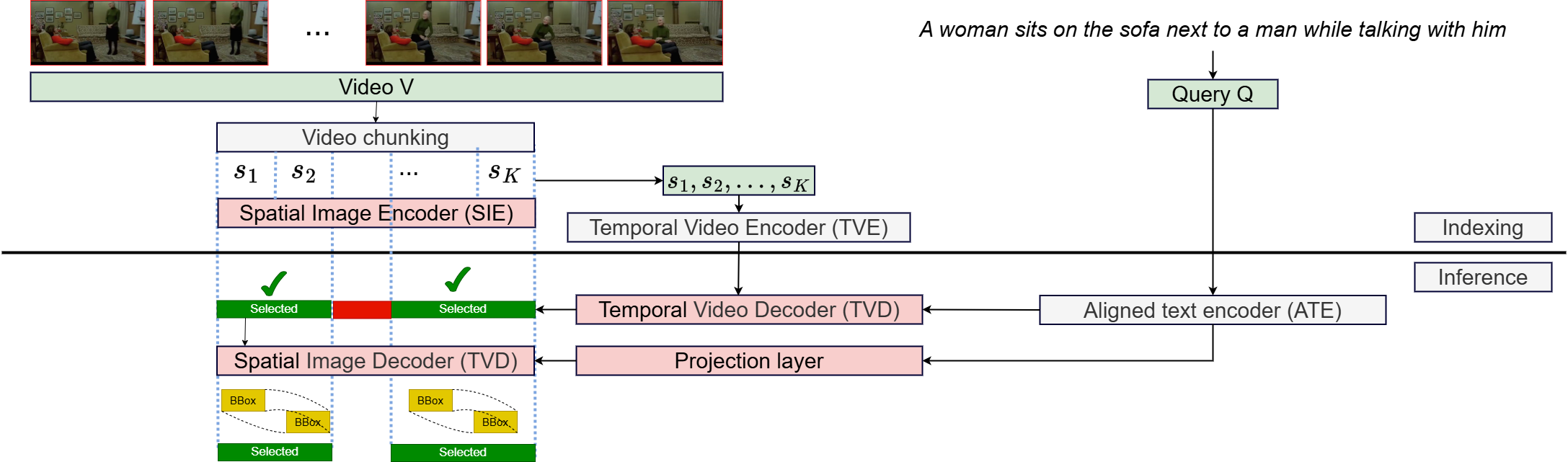}
    \caption{Blueprint of our proposed P-STVG architecture. The horizontal line separates indexing and inference phase, while red boxes represents modules we fine-tuned or trained, while gray boxes denote pre-trained modules that were kept frozen.  }
    \label{fig:blueprint}
\end{figure*}

\section{Related works}

\subsection{Fully supervised STVG}
In this setting, end-to-end neural networks are trained on fully annotated datasets with both spatial and temporal labels to predict spatio-temporal tubes. Early STVG methods generated candidate tubes before selecting the correct one \cite{zhang2020does,tang2021human,yamaguchi2017spatio}, whereas recent transformer-based approaches directly predict the target tube in a single-stage pipeline \cite{su2021stvgbert,yang2022tubedetr,jin2022embracing,wasim2024videogrounding,lin2023collaborative,gu2025knowing}.

Among them, \cite{yang2022tubedetr} first introduced a transformer architecture with a dual-stream encoder and a space-time decoder. This was extended by \cite{gu2024context}, which incorporated object visual cues for context guidance and a 3D encoder \cite{liu2022video} to better capture temporal information. More recently, \cite{gu2025knowing} introduced target-aware modules that generate object queries from video-text cues, improving performance but requiring prior target knowledge, typically provided as an additional word in the query, limiting practical applicability, while \cite{gu2026towards} exploited an autoregressive transformer for grounding a text-described person or event in long videos.
\cite{tu2026bridging} first identified when a queried event occurs, then used a separate query-guided spatial decoder to localize the referred object in each relevant frame. Its main contribution is a semantic bridge between temporal reasoning and precise bounding-box prediction in the spatial decoder.

Overall, these methods primarily optimize benchmark performance, often at the expense of model size and generalization.

\subsection{Weakly supervised STVG}

Weakly supervised STVG aims to localize spatio-temporal tubes using only video-level descriptions, avoiding costly frame-level bounding box annotations.

Although weak supervision has been widely studied for image grounding \cite{wang2020maf,liu2021relation,gupta2020contrastive} and video temporal grounding \cite{zheng2022weakly,chen2022weakly}, extending it to spatio-temporal grounding remains relatively unexplored. To mitigate the spurious correlations observed in STVG models, \cite{li2023winner} proposed a language decomposition tree with structural attention to align linguistic components and visual regions at multiple levels. \cite{jin2024weakly} instead formulates STVG as region-phrase and frame-sentence alignment, models ground-truth alignments as latent variables, and optimizes the joint distribution using variational EM. Addressing the limited temporal consistency of foundation models such as Grounding DINO \cite{liu2024grounding}, \cite{kumar2025contextual} employs GPT-3 to decompose complex queries, links textual queries into tubelets, and adopts a curriculum learning strategy from simple to complex videos. Similarly, \cite{garg2025stpro} bridges image foundation models and video grounding through a tubelet referral grounding module that connects frame-level boxes into tubelets, combined with curriculum learning from sparse actions to dense scenes, while \cite{li2026tubermc} generates text-conditioned object tubes and refines their spatial and temporal boundaries by reconstructing masked query phrases.

\subsection{Zero-shot STVG}

Zero-shot STVG seeks to localize spatio-temporal tubes without task-specific training or annotated bounding boxes and temporal intervals. Early approaches adapted pretrained vision-language models not designed for STVG: \cite{shtedritski2023does} guides CLIP attention by marking target objects with red circles, while \cite{subramanian2022reclip} combines cropped and blurred object candidates with a symbolic spatial resolver to improve relational reasoning. The first training-free zero-shot STVG method, \cite{bao2024e3m}, leverages pretrained models including CLIP, DINO, Grounding DINO, and SAM \cite{liu2024grounding,ravi2024sam}. It formulates temporal and spatial localization as latents, using an E-step to suppress background noise and an M-step with Kalman filtering and visual prototypes for spatial grounding.

 \cite{yang2025unleashing} exploits grounding tokens from multimodal large language models (MLLMs) \cite{li2024llava,chen2024sharegpt4video,wang2024qwen2}. More recently, agentic based frameworks have been proposed for STVG;  \cite{zhao2026agentic} exploits 
Spatial Reasoning Agent and Temporal Reasoning Agent to collaboratively propose, track, verify, and temporally trim an object tube matching a text query.

%% file: sec/3_method.tex
\section{Method}
\subsection{Preliminaries}
Given a video-query pair (V,Q), the goal is to localize the spatio-temporal tube $BB = \{ bb_{t}\}_{t = t_s}^{t_e}$ where $bb_t$ is the bounding box for time $t$ and ($t_s$,$t_e$) indicate the start and the end of the detected time interval, respectively. 
In our pipeline, we divide the STVG task into two cascaded sub-tasks: first, temporal grounding, where we identify the correct time interval, followed by spatial grounding, where we extract the bounding boxes.

 As shown in Fig. \ref{fig:blueprint}, our solution is structured into two distinct phases: \textbf{indexing} and \textbf{inference}. The former encompasses all offline steps that do not depend on a specific text query, including the extraction of video embeddings for both spatial and temporal detection, while the latter  involves the decoding process, where text embeddings are compared and processed to generate the final tube.

\subsection{General architecture} \label{general_atch}

During indexing, a video $V$ is divided into $K$  non-overlapping chunks of equal size, resulting in a partition $V = \{s_i\}_{i = 1}^{K}$, where $s_i$ stands for the $i$-th video chunk.
Each $s_i$ is then fed into the \emph{temporal video encoder} (TVE) obtaining $K$ time-aware semantic video embeddings $E^{v} = \{ e_i^{v}\}_{i = 1}^{K} = \{ TVE(s_i)\}_{i = 1}^{K}$.
In parallel, for each  $s_i$  we sample NF frames $\{f_{i}^1,...,f_{i}^{NF} \}$,  which are fed into the \emph{Spatial Image Encoder} (SIE) that processes them individually to determine spatial-aware image embeddings $E_{i}^{img} = \{ e_{ij}^{img}\}_{j = 1}^{NF} = \{ SIE(f_{i}^{j})\}_{j = 1}^{NF}$.

During the inference, we start with an input text query $Q$, which is fed into the \emph{aligned text encoder} (\emph{ATE}) to obtain the corresponding text embedding  $s_{q} = ATE(Q)$. Next, the embedding is fed into the \emph{temporal video decoding}  (\emph{TVD}) block along with the $E^v$ in order to predict the temporal interval $TI = (t_s,t_e)$.
Once we have $TI$, we can retrieve $E_{(t_s,t_e)}^{img}$ relative to the detected interval and fed them along with $s_q$ to the \emph{Spatial Image Decoder} (\emph{SID}) to obtain the final bounding boxes $BB = \{ bb_{t}\}_{t = t_s}^{t_e} = SID(s_q;E_{(t_s,t_e)}^{img} )$. Before $TVD$, $s_q$ is passed through a projection layer that helps alignment with the SID, which is fine-tuned as explained in Section~\ref{sec:tea}.

Importantly, in our setting we treat the temporal and spatial grounding as distinct tasks, allowing users to customize their focus and objectives, for example by prioritizing temporal grounding, spatial grounding, or both. 

\subsection{Temporal video grounding}
As part of the indexing stage, the video is divided into $K$  non-overlapping  equal-size chunks, obtaining $V = \{s_i\}_{i = 1}^{K}$, possibly achieving a reasonable trade-off between granularity and coherence of video content related information. 
We designed two distinct TVD techniques, to be used at inference stage. 
\subsubsection{\textbf{Temporal Video Encoder (TVE)}} To extract temporal-aware embeddings, we exploited pre-trained Tiny MobileViCLIP~\cite{yang2025mobileviclip}. This architecture is meticulously designed to ensure both reliability and efficiency, enabling easy deployment on edge devices such as smartphones. 
Specifically, building upon MobileCLIP \cite{vasu2024mobileclip}, it employs an enhanced hybrid vision transformer, MCi0, as its image encoder \cite{vasu2023mobileone}. For the text encoder, it utilizes an efficient hybrid model based on RepMixer \cite{vasu2023fastvit}.

With Tiny MobileViCLIP we obtained video embeddings $E^v = \{ e_i^v\}_{i = 1}^{K} = \{ TVE(s_i)\}_{i = 1}^{K}$, which are used with the text embedding $s_{q}$ to obtain temporal interval of interest at the inference stage. 

\begin{algorithm}[t]
\caption{Shot-level Temporal Grounding via Thresholding}
\label{alg:shot_grounding}
\begin{algorithmic}
\Require $E^{v} = \{e^{v}_{i}\}_{i=1}^{K}$, $s_{q}$, $\tau$
\State \textbf{Phase 1: Similarity computation}
\State $S^{vq}_{E^{v}} \gets \{CS(e^{v}_{i}, s_{q})\}_{i=1}^{K} \in \mathbb{R}^{K}$
\State $\hat{S}^{vq}_{E^{v}} \gets MinMax(S^{vq}_{E^{v}})$
\State \textbf{Phase 2: Thresholding}
\State $th \gets mean(\hat{S}^{vq}_{E^{v}}) - \tau \cdot std(\hat{S}^{vq}_{E^{v}})$ \Comment{Define Thresholding}
\State $\bar{\mathbf{v}} \gets zeros(K)$
\For{$i = 1$ \textbf{to} $K$}
    \If{$\hat{S}^{vq}_{E^{v}}[i] \geq th$}
        \State $\bar{\mathbf{v}}[i] \gets 1$
    \Else
        \State $\bar{\mathbf{v}}[i] \gets 0$
    \EndIf
\EndFor
\State \textbf{Phase 3: hole filling and time interval extraction}
\State $\bar{\mathbf{v}} \gets gaps\_fill(\bar{\mathbf{v}})$
\State $t_{s} \gets find\_first\_moment(\bar{\mathbf{v}})$
\State $t_{e} \gets find\_last\_moment(\bar{\mathbf{v}})$
\State \Return $TI = (t_{s}, t_{e})$
\end{algorithmic}
\end{algorithm}

\subsubsection{\textbf{Temporal Video Decoder with thresholding}}\label{threshold} In this scenario, we extract the temporal interval with an adaptive zero-shot thresholding technique. We first compute the similarity embedding $\hat{\mathrm{S}}_{E^v}$ as follows:
\begin{align}\label{eq:sim}
    \mathrm{S}^{vq}_{E^v} &= \{CS(e_i^v,s_{q}) \}_{i = 1}^{K} \in \mathbb{R}^{K},   \\
    \hat{\mathrm{S}}^{vq}_{E^v} &= MinMax(\mathrm{S}^{vq}_{E^v}), \notag
\end{align}
where $CS$ and $MinMax$ represent the cosine similarity function and Min-Max normalization, respectively. Once we have the $\hat{\mathrm{S}}_{E^v}$, which has values from 0 to 1, we first compute the adaptive threshold in the following way:
\begin{equation}
th = mean(\hat{S}^{vq}_{E^{v}}) - \tau \cdot std(\hat{S}^{vq}_{E^{v}}),
    \label{eq:threhold_th}
\end{equation}

where  $\tau$  is fixed to 0.5 a priori, without tuning on the evaluation sets, to preserve the fully zero-shot nature of this variant; the supplementary material reports a sensitivity analysis showing that performance is stable over a broad range of  $\tau$ . For each chunk, we determine its inclusion within a temporal interval based on whether its corresponding similarity score exceeds the predefined threshold, $th$, defined in eq. \ref{eq:threhold_th}, obtaining the binary vector $\bar{\mathbf{v}}$. Subsequently, we apply an automatic post-processing step to eliminate noisy gaps in the resulting binary sequence; specifically, we perform hole-filling on a 1D sequence and extract the largest connected component, discarding all other smaller segments. This ensures that only the primary signal interval is preserved while bridging minor internal gaps. Ultimately, the predicted temporal intervals $TI = (t_s,t_e)$ are derived by identifying the first and the last selected shots within the longest continuous interval. This procedure requires no human intervention and is described by Algorithm \ref{alg:shot_grounding}.


This approach offers two key benefits: (i) it operates in a zero-shot manner, as no part of the pipeline has been exposed to relevant datasets, and (ii) it avoids adding extra parameters, leveraging existing pretrained components' knowledge to address the task.
\begin{figure}[!ht]
    \centering
    \includegraphics[width=0.65\linewidth]{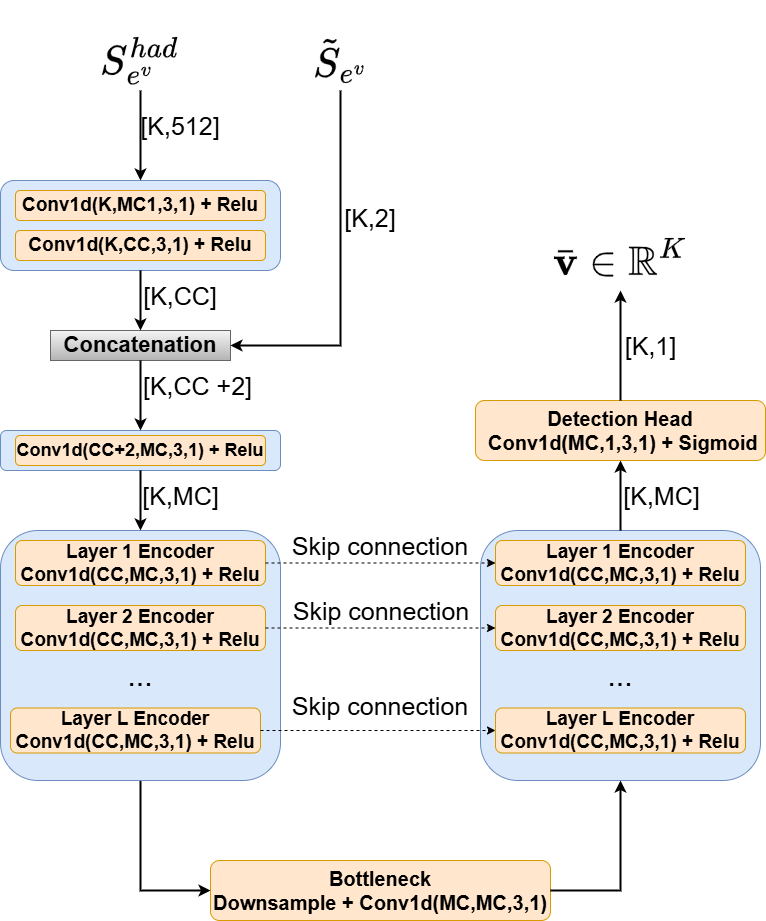}
    \caption{Diagram of the \textbf{T-Unet}. Values inside boxes are input dimension, output dimension, kernel size, and padding. On square brackets there is embedding dimensions at the specific stage.}
    \label{fig:unet}
\end{figure}

\begin{algorithm}[t]
\caption{Shot-level Temporal Grounding via T-Unet}
\label{alg:shot_grounding_unet}
\begin{algorithmic}
\Require $E^{v} = \{e^{v}_{i}\}_{i=1}^{K}$, $s_{q}$
\State \textbf{Phase 1: Input computation}
\State $i^{*} \gets \operatorname*{arg\,max}_{i=1,\dots,K} CS(e^{v}_{i}, s_{q})$
\State $e^{v}_{i_{max}} \gets e^{v}_{i^{*}}$
\State $S^{vq}_{E^{v}} \gets \{CS(e^{v}_{i}, s_{q})\}_{i=1}^{K} \in \mathbb{R}^{K}$
\State $S^{vv}_{E^{v}} \gets \{CS(e^{v}_{i}, e^{v}_{i_{max}})\}_{i=1}^{K} \in \mathbb{R}^{K}$
\State $\tilde{S}_{E^{v}} \gets [\,S^{vv}_{E^{v}} : S^{vq}_{E^{v}}\,]$ \Comment{Concatenation}
\State $S^{had}_{E^{v}} \gets \{Hadamard(e^{v}_{i}, s_{q})\}_{i=1}^{K} \in \mathbb{R}^{K \times D}$
\State \textbf{Phase 2: Thresholding with T-Unet}
\State $\bar{\mathbf{v}} \gets \text{T-Unet}(S^{had}_{E^{v}}, \tilde{S}_{E^{v}})$
\State \textbf{Phase 3: hole filling and time interval extraction}
\State $\bar{\mathbf{v}} \gets gaps\_fill(\bar{\mathbf{v}})$
\State $t_{s} \gets find\_first\_moment(\bar{\mathbf{v}})$
\State $t_{e} \gets find\_last\_moment(\bar{\mathbf{v}})$
\Return $TI = (t_{s}, t_{e})$
\end{algorithmic}
\end{algorithm}

\subsubsection{\textbf{Temporal video decoder  with Temporal Unet}} \label{t-unet} In this scenario, we introduce a lightweight module termed \textbf{T-Unet}, which employs a UNet-like architecture with one-dimensional convolutions, as illustrated in Fig. \ref{fig:unet}. 
The goal of this module is to combine information gathered by $E^v$, acting as an extractor of long-range temporal dependencies, enabling information aggregation between different video chunks.

we first extract the shot similarities $S^{vq}_{E^{v}}$ between $E^{v}$ and $s_{q}$,
then we define $e^{v}_{i_{max}}$ as follows:
\begin{equation}
i^{*} = \operatorname*{arg\,max}_{i=1,\dots,K} CS(e^{v}_{i}, s_{q}),
\qquad
e^{v}_{i_{max}} = e^{v}_{i^{*}},
\end{equation}
which is the embedding of the video chunk with the highest similarity with
respect to the query, used as the reference for the other embeddings.

We then concatenate the two similarity vectors obtaining $\tilde{\mathrm{S}}_{E^v} = [\mathrm{S}^{vv}_{E^v}:\mathrm{S}^{vq}_{E^v}]$, which constitutes the first input of our decoder.
Inspired by SimBase \cite{bao2024simbase}, the second input utilizes a straightforward \emph{Hadamard product} between chunks and query embeddings,
\begin{equation}
       \mathrm{S}^{had}_{E^v} = \{Hadamard(e_i^v,s_q) \}_{i = 1}^{K} \in \mathbb{R}^{K \times D},
\end{equation}
where $D$ is the embedding dimension.
This way we avoid complex language-video interactions and significantly simplify the architecture.
First, $\mathrm{S}^{had}_{E^v}$ is projected into a latent space of dimension  CC (compression channel) via an intermediate representation of dimension MC (middle channel). It is then concatenated with $\mathrm{S}^{vv}_{E^v}$ and projected into a latent space of dimension $MC$. Finally, resulting embedding is inputted to the \textbf{T-Unet} with $M$ encoder-decoder layers, generating the final binary response for each chunk, denoted as $\bar{\mathbf{v}} \in \mathbb{R}^{K}$.
The vector $\bar{\mathbf{v}}$ is passed through a sigmoid layer, constraining its values to the range [0, 1]. Following a procedure analogous to that described in \ref{threshold}, we apply a threshold operation, resulting in a binary vector where a value of 1 indicates detected shots and 0 signifies excluded ones; the same type of hole-filling post-processing is applied as in TVD with thresholding. The entire process is described in Algorithm~\ref{alg:shot_grounding_unet}.

\subsection{Spatial video grounding} \label{sec:tea}
Once the temporal video grounding is addressed, i.e. we have $TI = (t_s,t_e)$, we utilize  image embeddings $E_{(t_s,t_e)}^{img}$ computed at the indexing stage and extract final bounding boxes by means of the SID $BB = \{ bb_{t}\}_{t = t_s}^{t_e} = SID(s_q;E_{(t_s,t_e)}^{img} )$. 
In order to accomplish this task, we employ \emph{MDETR} \cite{kamath2021mdetr}, which is a well-known end-to-end integrated multimodal object detection model that identifies objects within images based on natural language queries.
However, \emph{MDETR} exploits RoBERTa \cite{liu2019roberta} text encoder, since it excels at understanding the context of words in a sentence, and it allows Soft Token Prediction that can associate span of text tokens with a detected box. 

To save parameters and computational resources, it would be optimal to use a single text encoder for both temporal and spatial grounding. This requires fine-tuning the MDETR architecture to align with the text encoder from MobileVICLIP, which is designed for cross-modal alignment rather than natural language understanding.

To perform cross-architecture synthesis, we adapt the MobileViCLIP text encoder to support character-level span prediction rather than global sequence representations.
In particular, we implemented an extension of the standard BPE tokenizer that generates a deterministic offset mapping $\mathcal{O} = \{(s_i, e_i)\}_{i=1}^N$, where each pair is the start and end character indices for the $i$-th subword token in the original query. 
Given a query description and its associated character span $[c_{start}, c_{end}]$, the model extracts the relevant token $\mathcal{T}$ in the following way:
\begin{equation}
    \mathcal{T} = \{ t_i \mid [s_i, e_i] \cap [c_{start}, c_{end}] \neq \emptyset \}
\end{equation}
As performed by MobileViCLIP, we generate a sequence of hidden states $H \in \mathbb{R}^{L \times D}$ where L is the context length and D is the feature dimension.
By exploiting character-level offsets, we are able to perform Soft Token Prediction, aligning the visual features extracted from the video backbone with specific sub-sequence of $H$.
Additionally, an attention mask $M \in \{0, 1\}^L$ is generated to differentiate valid semantic content from padding tokens. This mask is incorporated into the cross-modal transformer layers, ensuring visual features do not attend to null text embeddings, stabilizing the   training  of the multimodal reasoning heads.

\begin{table*}[!th]
\centering
\caption{Quantitative comparison on HCSTVG-v1/v2, between our methods and SOTA algorithms. Here, m\_t and  v\_t stand for m\_tIoU and m\_vIoU, respectively.  Light red indicates our method. STVG-supervised parameters count only modules exposed to STVG annotations }
\label{tab:results}
\small
\begin{tabular*}{\textwidth}{@{\extracolsep{\fill}} cccc|cccc|cccc @{}}
\toprule
\textbf{Sup.} & \textbf{Method} & \textbf{STVG-sup. Pars} & \textbf{ Total Pars.} & \multicolumn{4}{c|}{\textbf{HCSTVG-v1}} & \multicolumn{4}{c}{\textbf{\textbf{HCSTVG-v2}}} \\
& & \textbf{(M)} & \textbf{(M)} & m\_t &  m\_v & v@.3 & v@.5 & m\_t &  m\_v & v@.3 & v@.5 \\ 
\midrule
\multirow{2}{*}{\textbf{Full}} 

 & Bridge-STG \cite{tu2026bridging} & $\sim$250–400 & $\sim$8.98k & -  & -& - & - & 64.1 & 41.5 & 67.5 &  38.6 \\
 & TA-STVG \cite{gu2025knowing} & 206 & 234 & 53  & 39.1 & 63.1 & 36.8 & 60.4 & 40.2 & 65.8 &  36.7 \\
  & ART-STVG \cite{gu2026towards} & 207 & 235 & - & - & - & - & 59.2& 39.2 & 64.4 &  33.2 \\
 & TubeDETR \cite{yang2022tubedetr}   & 185  & 185 & 43.7  & 32.4 & 49.8 & 23.5 & 53.9   & 36.4 & 58.8 & 30.6  \\ 
& CG-STVG \cite{gu2024context}  & 203 & 231 & 52.8  & 38.4 & 61.5  & 36.3 & 60  & 39.5 & 64.5 & 36.3  \\ 
\midrule
\multirow{2}{*}{\textbf{Weak}} 
 & WINNER \cite{li2023winner}   & $\sim$22-30 & $\sim$162-275  & -  & 14.2 & 17.2  & 6.1 & - & - & -& - \\
 & CoSPal \cite{kumar2025contextual}    & $\sim$20-32 & $\sim$160-262 & -  & 22.2 & 31.4  & 18.9 & -  & 16.0 & 20.1 & 13.1 \\
 &  STPro \cite{garg2025stpro}   & $\sim$22-40 & $\sim$162-285 & -  & 17.6 & 27.0  & 12.9 & -  & 20.0 & 31.1 & 14.6 \\
  &  TubeRMC \cite{li2026tubermc}   & $\sim$40-100 & $\sim$230-300 & - &19.38 &23.88 & 6.75 & -  & - & - & - \\
\midrule
\multirow{2}{*}{\textbf{Zero-shot}} 
  & ASTG \cite{zhao2026agentic}   & 0 & $\sim$30-100k & -  & 32.3 & 54.4  & 28.2 & - & 34.8 & 55.5 & 31.4 \\
    & RealVG \cite{wei2025realvg}   & 0 & $\sim$7k & -  & 29.5 & 40.0  & 25.8 & - & 33.1 & 42.3 & 27 \\
  & Unleash \cite{yang2025unleashing}   & 0 & $\sim$7k  & -  & 24.8 & 41.5  & 16.3 & - & 27.7 & 44.7 & 19.5 \\
 & E3M \cite{bao2024e3m}     & 0 & $\sim$7.5k & -  & 19.1 & 29.4 & 10.6 & -  & - & - & - \\
  & RedCircle \cite{shtedritski2023does}     & 0 & $\sim$149-423 & -  & 9.2 & 7.8 & 1.6 & -  & - & - & - \\
    & ReClip \cite{subramanian2022reclip}     & 0 & $\sim$149-423 & -  & 14.4 & 18.3 & 4.9 & -  & - & - & - \\
\midrule
\rowcolor{rowred} \textbf{Zero-shot} &  \textbf{P-STVG} \tiny{Threshold}   & 0 & \textbf{82.28}  & 35.28  & 16.86 & 18.06  & 3.34& 45.19 & 22.31 & 27.15 & 6.82 \\
\midrule
\rowcolor{rowred} \textbf{Spatial zero-shot}  & \textbf{P-STVG} \tiny{T-Unet}  & 0.795 & \textbf{83.07} & 40.95 & 18.92  & 22.65 & 3.12  &  53.11  & 22.84 & 31.65 & 9.15  \\
\bottomrule
\end{tabular*}
\end{table*}

\subsection{Training strategy}

\subsubsection{\textbf{MDETR tuning with text encoder alignment}} \label{mdetr_training}

We fine-tune the MDETR architecture~\cite{kamath2021mdetr} to align with the MobileViCLIP text encoder, starting from an EfficientNet-based pretrained model~\cite{tan2019efficientnet} and following the original training strategy (soft token prediction and contrastive alignment) on Flickr30k~\cite{plummer2015flickr30k}, MS COCO~\cite{lin2014microsoft}, and Visual Genome~\cite{krishna2017visual}.
Inspired by~\cite{yang2024clip}, we add a distillation term encouraging our embeddings to match those of a frozen, RoBERTa-based MDETR teacher. Let $\mathbf{w}_{im}^{s}, \mathbf{w}_{im}^{t}$ and $\mathbf{w}_{q}^{s}, \mathbf{w}_{q}^{t}$ denote the student/teacher outputs of the image and text encoders, respectively:
\begin{equation} \label{eq_dist}
    \mathcal{L}_{distill} = \frac{\lambda}{2} \cdot \big( mse(\mathbf{w}_{im}^{s},\mathbf{w}_{im}^{t}) + mse(\mathbf{w}_{q}^{s},\mathbf{w}_{q}^{t}) \big).
\end{equation}
This term is added to the loss of~\cite{kamath2021mdetr} with $\lambda{=}1$; alignment of the text embedding is enabled by the trainable projection layer (Fig.~\ref{fig:blueprint}).

\subsubsection{\textbf{T-Unet training}}
The second step is to train \textbf{T-Unet} in order to be able to extrapolate the right time interval, as described in Section \ref{t-unet}.
In order to accomplish this task, we adopted the Dice loss function:
\begin{equation}
\mathcal{L}_{Dice} = 1 - \frac{2 \sum_{i=1}^{K} \hat{p}_i y_i + \epsilon}{\sum_{i=1}^{K} \hat{p}_i + \sum_{i=1}^{K} y_i + \epsilon},
\end{equation}
where $K$ is the number of shots, $\hat{p}_i \in [0, 1]$ is the prediction for the $i$-th shot, $y_i \in \{0, 1\}$ is the ground truth, and $\epsilon$ is a small constant to ensure stability. 
We trained this architecture for 50 epochs, using \num{1e-4} as learning rate with Adam \cite{kingma2014adam}, reducing it after $35$ epochs to \num{5e-5}.





%% file: sec/4_exp.tex
\section{Experiments}

\subsection{\textbf{Datasets and Evaluation}}
We evaluate on HCSTVG-v1/v2 \cite{tang2021human} and VidSTG
\cite{zhang2020does}. HCSTVG-v1 contains 5,660 video-text pairs; HCSTVG-v2
extends it to 10,131 training videos and 2,000 test samples. VidSTG
comprises over 90k video-text pairs with declarative and interrogative
sentences, of which we use only the declarative validation set.

Following \cite{kumar2025contextual, yang2022tubedetr}, we report
\emph{m\_tIoU}, \emph{m\_vIoU}, and \emph{vIoU@R}. \emph{m\_tIoU} is the
mean temporal IoU between predicted and ground-truth intervals, measuring
temporal localisation alone. For a single sample, denoting by $S_i$ and
$S_u$ the intersection and union of the predicted and ground-truth
intervals, $vIoU = \frac{1}{S_u} \sum_{t \in S_i} IoU(\hat{b}_t, b_t)$,
where $\hat{b}_t$ and $b_t$ are the predicted and ground-truth boxes at
frame $t$; \emph{m\_vIoU} is its average over the dataset. Finally,
\emph{vIoU@R} is the fraction of samples with $vIoU > R$, with
$R \in \{0.3, 0.5\}$.

\subsection{\textbf{Configuration}}
Videos are divided into 20 non-overlapping chunks for HCSTVG-v1/v2 and 32
for VidSTG (roughly one embedding per second), balancing temporal decoder
granularity, sufficient frame-level variability for MobileViCLIP, and a
manageable number of chunks at indexing time. For \textbf{T-Unet} we fix
$L=4$, $MC=128$, and $CC=8$, giving a $\sim$0.795M-parameter network
trained only on temporal labels, ignoring bounding boxes and leaving the
rest of the pipeline unchanged.

Our method is spatially zero-shot with T-Unet and fully zero-shot with thresholding: no component is ever trained on STVG data. The spatial branch is fine-tuned only on image-grounding datasets and the T-Unet only on temporal labels, so the parameters we report as STVG-supervised in Tab.\ref{tab:results} count only the latter. We report total parameters alongside, so that the two accounting conventions can be compared directly.

\subsection{Comparison with state-of-the-art methods}
\subsubsection{\textbf{Quantitative results}} \label{quant_res}

Table \ref{tab:results} compares our method against fifteen
\textit{state-of-the-art} (SOTA) methods: five fully supervised
\cite{gu2025knowing,yang2022tubedetr,gu2024context,tu2026bridging,gu2026towards},
four weakly supervised
\cite{li2023winner,garg2025stpro,kumar2025contextual,li2026tubermc}, and six
zero-shot
\cite{yang2025unleashing,bao2024e3m,shtedritski2023does,subramanian2022reclip,wei2025realvg,zhao2026agentic}.
We also report trainable and total parameters, estimated when not stated in
the original paper (details in the supplementary material).

Against weakly supervised methods the picture depends on the dataset. On
HCSTVG-v2, both variants exceed all such methods  in terms of
m\_vIoU (22.31 and 22.84 against 20.0 for \emph{STPro} and 16.0 for
\emph{CoSPal}), despite using no Spatial annotation. On HCSTVG-v1 the margin
narrows: $\textbf{P-STVG}_{Unet}$ surpasses \emph{WINNER} and \emph{STPro}
(18.92 against 14.2 and 17.6) but trails \emph{TubeRMC} (19.38) and
\emph{CoSPal} (22.2), while the thresholding variant (16.86) exceeds only
\emph{WINNER}. On both datasets our v@.5 remains below these methods,
indicating that the gap is concentrated in strict spatial overlap rather
than in the coarser localization captured by m\_vIoU. 
These results are
obtained with a smaller parameter budget and a single forward pass, split
into indexing and inference, rather than the multi-stage training and
inference schemes these methods usually employ.

To locate the source of this spatial gap, we evaluate the spatial branch
with ground-truth temporal intervals. Under oracle temporal localization,
P-STVG reaches 39.73 m\_vIoU, 62.45 v@.3, and 32.85 v@.5 on HCSTVG-v2,
comparable to fully supervised methods such as \emph{TA-STVG} (40.2
m\_vIoU, 36.7 v@.5) and \emph{ART-STVG} (39.2, 33.2) despite using MDETR as off-the-shelf grounding model.
Relative to our end-to-end results (22.84 m\_vIoU, 9.15
v@.5), this shows that the spatial decoder is accurate and that most of
the loss comes from imprecise temporal intervals. 

Among zero-shot methods, P-STVG clearly outperforms \emph{RedCircle} and
\emph{ReCLIP}, while agentic and MLLM-based approaches (\emph{ASTG},
\emph{RealVG}, \emph{Unleash}) achieve higher accuracy using models of
several billion parameters and considerably more complex inference.  

On VidSTG (Tab. \ref{tab:vistg}) the same pattern holds. In the spatial
zero-shot setting, $\textbf{P-STVG}_{Unet}$ attains the best temporal
accuracy among non-fully-supervised methods (m\_tIoU 46.57 against 41.10
for \emph{CoSPal}) and an m\_vIoU comparable to weakly supervised
approaches (15.70 against 15.5--16.0), while remaining below them on v@.5.
The fully zero-shot variant is weaker (m\_vIoU 14.25), exceeding only
\emph{RedCircle}. 
Throughout, our full architecture stays below 90M
parameters, nearly two orders of magnitude smaller than the MLLM-based
zero-shot methods, placing P-STVG at a distinct operating point defined by
deployment feasibility rather than maximum accuracy.

\begin{table}[ht!]
    \centering
    \caption{Spatio-temporal performance on VidSTG declarative.}
    \label{tab:vistg}
    \begin{tabular}{l c c c c }
        \toprule
        \textbf{Method} & \textbf{m\_tIoU} & \textbf{m\_vIoU}  & \textbf{v@.3} & \textbf{v@.5} \\
        \midrule
        Bridge-STG & 52.6 & 37.2 & 52.4 & 37.4 \\
        TA-STVG & 51.7 & 34.4 & 48.2 & 33.5 \\
        \toprule
        STPro & - & 15.5& 19.4 &12.7 \\
        COSPAL & 41.10 & 16 & 20.1 &13.1 \\
        Tube-RMC & - & 15.93& 25.16 & 9.09 \\
        \toprule
        RedCircle & - & 8.6 & 7.6 & 0.9 \\
        E3M & - & 16.2 & 20.5 & 11.9 \\
        Unleash & - & 18 & 29.8 & 12.2 \\
        RealVG & - & 29 & 35.0 & 25.5 \\
        \toprule
        \toprule
        \rowcolor{rowred}
        P-STVG \tiny{Threshold} & 33.47 & 14.25  & 16& 3.4  \\
        \rowcolor{rowred}
        P-STVG \tiny{T-Unet} & 46.57 & 15.70  & 19 & 9.16  \\
        \bottomrule
    \end{tabular}
\end{table}

\begin{table}[h!]
    \centering
    \caption{Efficiency analysis on HC-STVG. We report peak VRAM, query time, and TFLOPs under the same inference protocol on an RTX 4060 GPU with 8 GB of VRAM. Blue indicates methods that could not be executed within this hardware budget and were measured on an H100 instead.}
    
    \label{tab:efficiency}
    \resizebox{\columnwidth}{!}{%
    \begin{tabular}{cccc}
        \toprule
         Method & \textbf{Peak VRAM (GB)} &
        \textbf{Query Time (s)} & \textbf{TFLOPs} \\
        \midrule
        TA-STVG & 4.36 & 0.90 & 2.87 \\
        \rowcolor{rowblue} UNLEASH & 14.71 & 3.35 & 8.09 \\
        \rowcolor{rowred} P-STVG \textsubscript{T-Unet}
        & \textbf{3.94} & \textbf{0.71} & \textbf{1.11} \\
        \bottomrule
    \end{tabular}%
    }
\end{table}

\begin{table*}[!t]
\centering
\caption{Results in terms of m\_tIoU,tIoU@3, and tIoU@5 on HCSTVG-v2 considering different \textbf{T-Unet} configurations. In light red the one chosen for the main architecture}
\vspace{-0.1cm}
\label{tab:tunet_config}
\small
\begin{tabular*}{\textwidth}{@{\extracolsep{\fill}} c|cccc|cc | ccc  @{}}
\toprule
 & \multicolumn{4}{c|}{\textbf{architecture configuration}} & \multicolumn{2}{c|}{\textbf{training details}} &\multicolumn{3}{c}{\textbf{performance}}  \\
ID & MC &CC & L & Total pars &  LR & BS & m\_tIoU & tIoU@3 &  tIoU@5  \\ 
\midrule
$TVD_1$ & 128 &8 &1 &351625 & 1e-4 &16 & 50.11 &78.45 &54.95 \\
$TVD_2$ & 512 &8 &1 &4749833 &1e-4 &16 &51.66 &80.75 &58.10 \\
$TVD_3$ & 512 &8 &2 &5536265 &5e-5 &16 &52.14 &81.75 &59.95 \\
$TVD_4$ & 512 &8 &3 &7896585 &5e-5 &16 & 52.47 &  81.75&61.60 \\
$TVD_{5}$ & 512 &8 &4 &10256905  &1e-4 & 16& 53.12 & 82.65 & 61.15 \\
\toprule
\rowcolor{rowred} $\mathbf{TVD^{*}}$ & \textbf{128} &\textbf{8} & \textbf{4} & \textbf{794961} & 1e-4  & 16 & 53.11 & 81.90 & 61.75 \\
\bottomrule
\end{tabular*}
\end{table*}

\subsubsection{Efficiency analysis}\label{comp_eff}

Tab. \ref{tab:efficiency} reports peak VRAM, query time, and TFLOPs at
inference, measured for a single text query on a 10-video subset of the
HC-STVG validation set using an RTX 4060 with 8\,GB of VRAM; the offline
indexing stage is reported separately for our method (tab. \ref{tab:number_shots}).
The subset is representative, as the videos are of comparable complexity. 
We report the
\textbf{T-Unet} variant only, since the difference with respect to
thresholding is negligible.  
We compare against TA-STVG and UNLEASH, representing strong supervised and zero-shot baselines, respectively. Because UNLEASH could not be executed within the 8 GB VRAM constraint and required an H100 GPU instead, its results are shown for reference but should be interpreted under a different hardware budget. Under this protocol, P-STVG achieves lower peak memory usage, shorter query time, and fewer TFLOPs,  highlighting the efficiency advantage of the proposed pipeline.

\subsection{Ablation study and Internal analysis}
\subsubsection{\textbf{Different T-Unet configuration}}
\begin{figure}[!ht]
    
    \centering
    \includegraphics[width=1\columnwidth]{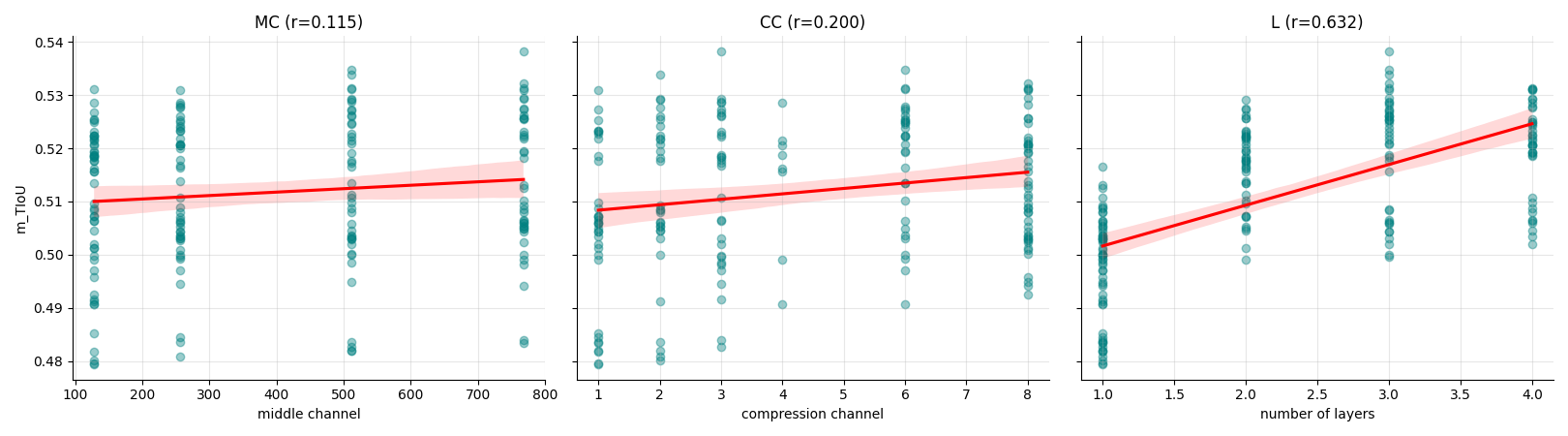}
    \vspace{-0.35cm}
    \caption{\emph{m\_tIoU} with respect to MC, CC, and L. Each blue dot is
one trained configuration from the full grid search. The red line is the
linear trend line that best fits them, the light red zone represents the
95\% confidence interval around the regression line, and $r$ is the
Pearson correlation.}

    \label{fig:correlation}
\end{figure}

The \textbf{T-Unet} configuration governs the trade-off between
\emph{m\_tIoU} and memory footprint.
Tab. \ref{tab:tunet_config} reports the most representative configurations
of \emph{compression channels} (CC), \emph{middle channels} (MC), and
number of \emph{layers} ($L$), together with \emph{learning rate} (LR) and
\emph{batch size} (BS), out of the full grid search summarised in Fig.
\ref{fig:correlation} (blued dots).
 Performance tracks architecture rather than
parameter count: $TVD^{*}$ outperforms $TVD_3$ and $TVD_4$ despite having
roughly an order of magnitude fewer parameters, and matches $TVD_5$ with
one thirteenth of its budget. Over the full sweep (Fig.
\ref{fig:correlation}), the Pearson correlation between \emph{m\_tIoU} and
$L$ is 0.62, against 0.11 and 0.20 for MC and CC, indicating that depth
contributes considerably more than channel width.

\begin{table}[ht]
    \centering
    \caption{Evaluation of our pipeline in terms of \text{m\_tIoU} considering different number of chunks (NC), on hcstvg-v2. We also report memory footprint for a single video (MF) and time for indexing (IT). Light red indicates main architecture configuration. }
    \label{tab:number_shots}
    \vspace{-0.35cm}
    \begin{tabular}{l  l l l  l}
        \toprule

        \textbf{NC} & \textbf{m\_tIoU} \tiny{\textbf{T-Unet}} & \textbf{m\_tIoU} \tiny{\textbf{Thresholding}}  &  \textbf{MF} (Kb) & \textbf{IT} (s) \\ \hline
        40  & 51.87 & 43.71 & 82 & 3.29 \\
        24  & 52.40 & 44.31 & 49.3  & 1.95 \\
        16  & 48.56 & 44.01 &  32.9 & 1.29 \\
        12  & 45.67& 43.83&  24.7 & 0.99\\
        \toprule
         \rowcolor{rowred} 20  & 53.11 &   45.19 & 44 & 1.61 \\
        \bottomrule
    \end{tabular}
\end{table}

\subsubsection{\textbf{Different number of chunks}}
Tab.\ref{tab:number_shots} reports effect of varying number of chunks when dividing the video. Increasing the number of chunks enlarges the video embedding (e.g., 49.3 KB for 24 chunks vs. 44 KB for 20) and increases indexing time due to the extraction of $E^{v}$ by TVE. However, performance does not improve monotonically, likely because of dataset characteristics, increased noise with excessive chunks, and MobileViCLIP requiring a sufficient temporal span to produce meaningful embeddings. Conversely, too few chunks (e.g., 12) reduce performance because of coarser temporal predictions.

%% file: sec/5_concl.tex
\section{Conclusion}

We presented Pocket-STVG, a lightweight framework for spatio-temporal video grounding that assembles efficient pre-trained components into a compact cascade split into an indexing and an inference stage. It comprises a temporal video encoder and aligned text encoder based on MobileViCLIP, a spatial encoder-decoder derived from MDETR, and a temporal video decoder implemented either as simple thresholding or as a lightweight U-Net. Video representations are computed independently of the query, so inference reduces to a single forward pass.

With fewer than 90M parameters, P-STVG matches weakly supervised methods in terms of m\_vIoU on HCSTVG-v2 and improves on earlier zero-shot approaches, at a fraction of their memory and computational cost.
Some limitations should be noted. Under strict spatial overlap our vIoU@0.5 remains below weakly supervised methods on both benchmarks, so the compact pipeline comes at a measurable accuracy cost when precise boxes are required. Furthermore, efficiency is measured on a resource-constrained desktop GPU rather than on mobile hardware

Improving temporal precision is therefore the most promising next step, through finer or overlapping chunking and by propagating boxes across frames to stabilize predictions near interval boundaries. The residual spatial margin could be narrowed by replacing per-frame detection with a video-based decoder that exploits temporal continuity, or by adapting the current one through parameter-efficient tuning such as LoRA. Validation on mobile hardware and on longer videos, where query-independent indexing should pay off most, is a natural further direction.

%% file: sec/7_appendix.tex
\maketitlesupplementary

\section{Distillation in MDETR tuning} 
On table \ref{tab:distillation} we have results on Flickr test dataset \cite{plummer2015flickr30k}, in terms of \emph{recall@k} with k = \{ 1,5,10\}, i.e. the percentage of queries for which at least one relevant item is found in the top-k results ranked by the model.
Light-blue row refers to the results from the teacher network, which is the EfficientNet-B3 based model trained by \cite{kamath2021mdetr}. As it is possible to observe, the major gain in our results happens when we fine-tune starting from the teacher itself, meaning that a good starting point is vital to align the model to a new text encoder. 
On the other hand, the contribution that the distillation term gave is marginal, even if present to a certain level; this suggests that this can be a promising direction to follow in future developments.

\begin{table}[h]
    \centering
    \caption{Results of MDETR fine-tuning for text encoder alignment in terms of Recall@\{1,5,10\}. Light blue line represents performance of the teacher.}
    \label{tab:distillation}
    \begin{tabular}{l l l l l}
        \toprule
        \textbf{Starting Net} & $\lambda$ & \textbf{R@1} & \textbf{R@5} & \textbf{R@10} \\ \hline
        Scratch  & 1 & 72.27 & 86.11 & 90.84  \\
        Scratch  & 0 & 72.09 & 85.55& 89.66\\
        \hline
        Teacher  & 1 & \textbf{79.41} & \textbf{91.84}& \textbf{94.48} \\
        Teacher  & 0 & 79.28 &  91.65&  94.29 \\
        \hline
        \rowcolor{rowblue} -  & - & \textbf{82.9}& \textbf{93.8} & \textbf{95.6} \\
        \bottomrule
    \end{tabular}
\end{table}

\section{Qualitative samples}
\begin{figure*}
    \centering
    \includegraphics[width=1\linewidth]{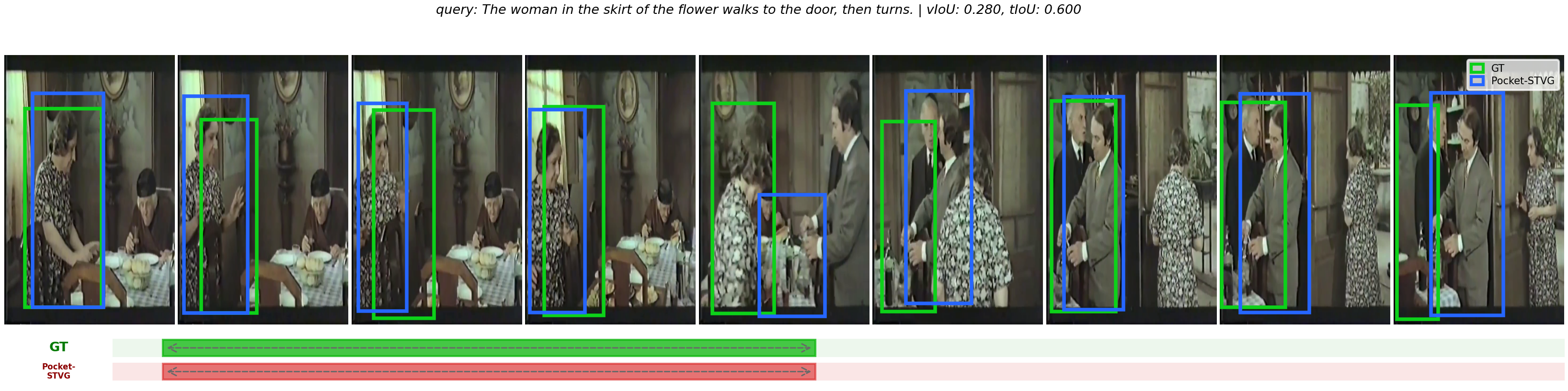}
    \includegraphics[width=1\linewidth]{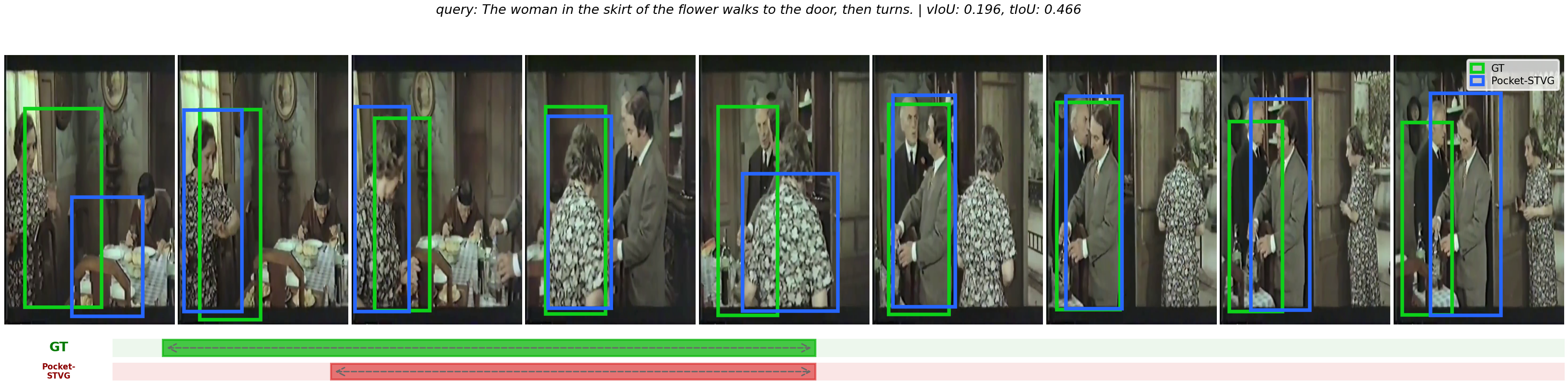}
    \caption{Qualitative samples taken from HCSTV-v2 using our method with thresholding temporal grounding (top) and Unet temporal grounding (bottom).  }
    \label{fig_qual_samples_1}
\end{figure*}

\begin{figure*}
    \centering
    \includegraphics[width=1\linewidth]{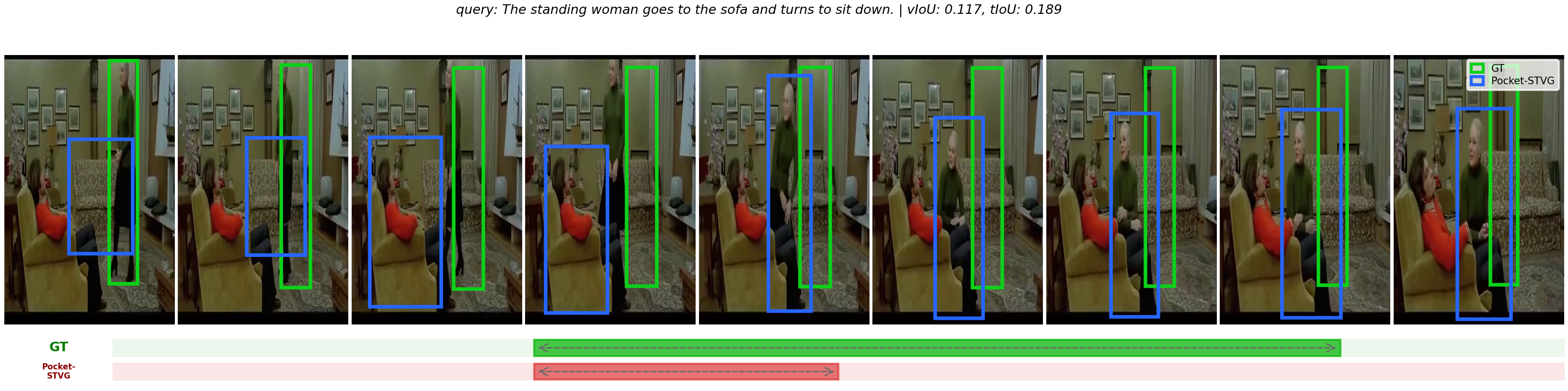}
    \includegraphics[width=1\linewidth]{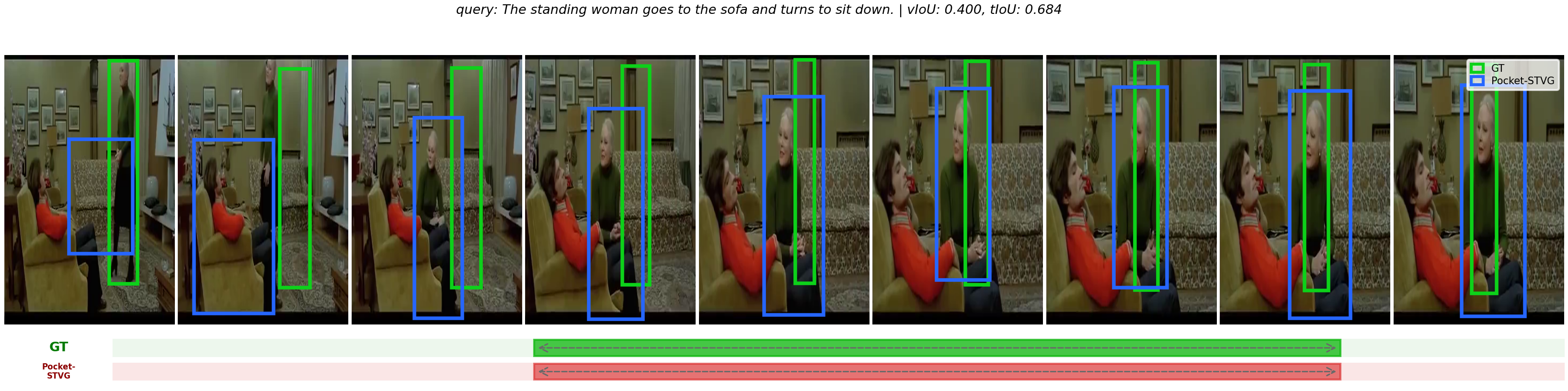}
    \caption{Qualitative samples taken from HCSTV-v2 using our method with thresholding temporal grounding (top) and Unet temporal grounding (bottom).  }
    \label{fig_qual_samples_2}
\end{figure*}

Fig. \ref{fig_qual_samples_1} presents qualitative examples in which thresholding outperforms U-Net temporal grounding. Note that the spatial grounding may differ even when the same method is applied, as the frames considered in each case are different.
Fig. \ref{fig_qual_samples_2} shows the same thing, but for a sample where our temporal grounding method outperformed plain thresholding.

\section{ Zero-shot quantitative results using different threshold}
In tab. \ref{tab:different_threshold} we analyze results in terms of \emph{m\_tIoU} considering our proposed method in a zero-shot configuration, i.e. by applying thresholding with different $\tau$. 
We consider HC-STVG v2 dataset and we just shot temporal results, since we are focusing only on temporal decoder module. 
We can observe that best results have been obtained  when the threshold is greater than 0.35.

\begin{table}[h]
    \centering
    \caption{Performance on HCSTVG-v2 considering our zero-shot methods with different threshold. Light blue indicates configuration chosen for the main paper. }
    \label{tab:different_threshold}
    \begin{tabular}{ l l l l}
        \toprule
         threshold $\tau$ & m\_tIoU & tIoU@3 &  tIoU@5  \\ \hline
         0.65 & 45.44 & 71.55 & 44.55 \\
         0.60 & 45.56 & 71.55 & 44.55 \\
         0.55 & 45.57 &  70.9 & 46 \\
         \rowcolor{rowblue} 0.5 & 45.19 & 70.15 & 46.2 \\
         0.45 & 44.81 & 69.65& 45.75\\
          0.4 & 44.76 & 69.5& 44.7 \\
         0.35 & 44.18 &  68.5&  44.7\\
         0.3 & 43.8& 67.95 & 44.45 \\
         0.25 & 43.59& 67.2 & 43.95 \\
        0.2 & 42.9& 65.85 & 43.6 \\
        0.15 & 42.6& 65.4 &  43.6 \\
        0.10 &  42.24& 64.6 & 43.6 \\
        \bottomrule
    \end{tabular}
\end{table}


\section{Estimation of state-of-the-art parameters}
In tab. \ref{tab:parameter_analysis} we show our parameters estimation for some SOTA methods. We highlight the fact that this estimation can be not exact, but it is has still a good level of precision useful to capture the order of magnitude of each method.

\begin{table*}[t!]
\centering
\caption{Parameter analysis of state-of-the-art method for zero-shot and weak-shot STVG.}
\label{tab:parameter_analysis}
\begin{tabular}{@{}lllcc@{}}
\toprule
\textbf{Method} & \textbf{Component} &  \textbf{Model used}& \textbf{Parameters} & \textbf{State} \\ \midrule
\multirow{3}{*}{Unleash \cite{yang2025unleashing}}  & Vision Encoder & CLIP-ViT-L/14 & $\sim$300M & Frozen\\
 & Language Model & LLaVA-v1.5-7B & $\sim$6.7B & Frozen \\
 & Projection Layer & MLP Adapter & $\sim$20M & Frozen\\ 
  & \textbf{TOTAL} &  -  & $\sim$\textbf{7B} & Frozen\\\midrule
 \multirow{3}{*}{E3M \cite{bao2024e3m}} & Vision Encoder & CLIP-ViT-L/14 & $\sim$300M & Frozen\\
 & Language Model & Video-LLaVA-7B & $\sim$7B & Frozen \\
 & Projection Layer & MLP Adapter & $\sim$20M & Frozen\\ 
  & Object refiner & Grounding DINO & $\sim$172M & Frozen\\ 
    & \textbf{TOTAL} &  -  & $\sim$\textbf{7.5B} & Frozen\\\midrule
 \multirow{3}{*}{RedCircle \cite{shtedritski2023does}} & Vision Encoder & CLIP (Vit-B/16 orViT-L/14) & $\sim$86-300M & Frozen\\
 & Text encoder & CLIP & $\sim$63-123M & Frozen \\
     & \textbf{TOTAL} &  -  & $\sim$\textbf{149-423}M & Frozen\\\midrule
 \multirow{3}{*}{ReCLIP \cite{subramanian2022reclip} } & Vision Encoder & CLIP (Vit-B/16 orViT-L/14) & $\sim$86-300M & Frozen\\
 & Text encoder & CLIP & $\sim$63-123M & Frozen \\
      & \textbf{TOTAL} &  -  & $\sim$\textbf{149-423}M & Frozen\\\midrule
 \multirow{3}{*}{WINNER \cite{li2023winner}} & Vision Encoder & Video-Swin / SlowFast & $\sim$30-120M & Frozen\\
 & Text encoder & RoBERTa / BERT & $\sim$110-125M & Frozen \\
  &  Hierarchical Decomposition & Action-Object decoder & $\sim$15M & Trainable \\
    &  Alignment Module & Spatio-Temporal Head & $\sim$5-10M & Trainable \\
&  Refinement Module & BBox Regression Head & $\sim$2-5M & Trainable \\
      & \textbf{TOTAL} &  -  & $\sim$\textbf{162-275}M & Frozen\\\midrule
     \multirow{3}{*}{CoSPal \cite{kumar2025contextual} } & Vision Encoder & Video-Swin / C3D & $\sim$30-120M & Frozen\\
     & Text encoder & BERT & $\sim$110M & Frozen \\
  &  Contextual Scoring & Self-placed Head & $\sim$10-15M & Trainable \\
    &  Interaction Module & Cross-Modal Attn & $\sim$8-12M & Trainable \\
    &  BBox Predictor & Regression Head& $\sim$2-5M & Trainable \\
      & \textbf{TOTAL} &  -  & $\sim$\textbf{160-262}M & Frozen\\\midrule
     \multirow{3}{*}{STPro \cite{garg2025stpro}} & Vision Encoder & Video-Swin / I3D & $\sim$30-120M & Frozen\\
     & Text encoder & BERT  / RoBERTa& $\sim$110-125M & Frozen \\
  &  Spatial Selector & Multiple Instance Learning & $\sim$8-15M & Trainable \\
    &  Temporal Module & Temporal decoder  & $\sim$12-20M & Trainable \\
    &  BBox Predictor & MLP Head& $\sim$2-5M & Trainable \\
      & \textbf{TOTAL} &  -  & $\sim$\textbf{162-285}M & Frozen\\\midrule
\multirow{6}{*} {ASTG~\cite{zhao2026agentic}}
 & Spatial Reasoning Agent   & Qwen3-VL-Plus (MLLM)      &  ---       & Frozen \\
 & Temporal Reasoning Agent  & Qwen3-VL-Plus (MLLM)      & ---       & Frozen \\
 & Query Parser (Controller) & Qwen3-Plus (LLM)          & ---        & Frozen \\
 & Object Tracker            & SAM2 (Hiera-L)            & $\sim$224M                 & Frozen \\
 & \textbf{TOTAL}                     & --                        & \textbf{$\geq\sim30B$} & Frozen \\
\midrule

\multirow{7}{*}{Bridge-STG~\cite{tu2026bridging}}
 & Vision Encoder      & Qwen3-VL ViT + patch merger        & $\sim$0.54B  & Frozen \\
 & Language Model      & Qwen3-8B (Qwen3-VL 7B)             & $\sim$8.2B   & Frozen \\
 & LoRA Adapters       & $r=8$, $\alpha=32$ (language tower)& $\sim$20M    & Trainable \\
 & Bridging Queries    & STSB ($M=8$ queries + MLP)         & $\sim$1M     & Trainable \\
 & Spatial Backbone    & Swin-L (COCO pre-trained)          & $\sim$197M   & Trainable \\
 & Spatial Enc-Decoder & QGSL (Deformable DETR, 6+6 layers) & $\sim$60M    & Trainable \\
 & TOTAL               & --                                 & $\sim$9.0B   & Trainable \\

 \bottomrule
\end{tabular}
\end{table*}

